# Adversarial Robustness and Feature Impact Analysis for Driver Drowsiness Detection

João Vitorino[1][0000-0002-4968-3653], Lourenço Rodrigues[2][0000-0002-2337-6911], Eva Maia[1][0000-0002-8075-531X], Isabel Praça[1][0000-0002-2519-9859] and André Lourenço[2][0000-0001-8935-9578]

[1] Research Group on Intelligent Engineering and Computing for Advanced Innovation and Development (GECAD), School of Engineering, Polytechnic of Porto (ISEP/IPP), 4249-015 Porto, Portugal
`{jpmvo,egm,icp}@isep.ipp.pt`
[2] CardioID Technologies, 1959-007 Lisboa, Portugal
`{lar,arl}@cardio-id.com`

**Abstract.** Drowsy driving is a major cause of road accidents, but drivers are dismissive of the impact that fatigue can have on their reaction times. To detect drowsiness before any impairment occurs, a promising strategy is using Machine Learning (ML) to monitor Heart Rate Variability (HRV) signals. This work presents multiple experiments with different HRV time windows and ML models, a feature impact analysis using Shapley Additive Explanations (SHAP), and an adversarial robustness analysis to assess their reliability when processing faulty input data and perturbed HRV signals. The most reliable model was Extreme Gradient Boosting (XGB) and the optimal time window had between 120 and 150 seconds. Furthermore, SHAP enabled the selection of the 18 most impactful features and the training of new smaller models that achieved a performance as good as the initial ones. Despite the susceptibility of all models to adversarial attacks, adversarial training enabled them to preserve significantly higher results, especially XGB. Therefore, ML models can significantly benefit from realistic adversarial training to provide a more robust driver drowsiness detection.

## 1 Introduction

According to the European Road Safety Observatory, 15 to 20% of all road crashes are estimated to be related to drowsy driving. In road accidents involving trucks and large vehicles, and especially those resulting in fatalities, this share is even higher [1]. While drivers have been sensibilized to the dangers of alcohol consumption or cellphone usage while driving, they are still, for the most part, dismissive of the impact that fatigue and drowsiness can have on their driving abilities [2].

It is essential to monitor drivers and provide them with timely warnings that their drowsiness level is excessive, prompting them to take protective measures. Furthermore, alerting drivers can improve awareness and accountability, or at least provide information for authorities to measure the real impact of fatigue and drowsiness on the road. Most currently available monitoring mechanisms rely on driving performance metrics like lane position and steering wheel dynamics, as these are easy to gather from vehicles [3]. However, they are only able to detect changes in driving behavior, so a warning is only provided after a driver starts demonstrating impairment [4].

To detect anomalous alterations before any behavioral impairment occurs, a solution can be to infer drowsiness levels directly from a driver's physiological signals like the Heart Rate (HR). Using HR Variability (HRV) as the source for the detection enables it to be based on the dynamics of cardiac rhythm modulation, consisting of the time intervals between each heartbeat. As cardiac rhythm is modulated by the balance of sympathetic and parasympathetic nervous systems, and that balance reflects the alertness state, it can provide valuable information for drowsiness detection [5].

This work presents multiple experiments with HRV time windows and Machine Learning (ML) models, followed by a feature impact and adversarial robustness analysis to assess their reliability when processing faulty input data and adversarial examples. Both regular and adversarial training approaches were compared to provide more robust ML models for driver drowsiness detection. This paper is organized into multiple sections. Section 2 provides a survey of previous work on driver HRV monitoring. Section 3 describes the utilized dataset, models, and methodologies. Section 4 presents and discusses the obtained results. Finally, Section 5 addresses the main conclusions.

## 2    Related Work

The application of HRV for driver drowsiness detection has yet to permeate the automotive market. Until recently, it was mostly hindered by the intrusive nature of Electrocardiogram (ECG) electrodes and the need for specialized knowledge to apply them and read a correct signal [6]. Nonetheless, the rise of consumer electronics like smart watches and bands equipped with Photoplethysmography (PPG) sensors, as well as off-the-person ECG recording devices like the CardioWheel [7], greatly reduced the knowledge barrier and allowed drivers to seamlessly gather physiological data.

Preparing for the future of interconnected physiological sensors, research efforts have been made on their usage with different features and time windows. In the Sleep-EYE [8] project, subjects drove in a Swedish highway while recording ECG and registering Karolinska Sleepiness Scale (KSS) levels, a 9-point self-assessment scale [9]. From the recordings, ML models were trained with HRV and Electrooculogram features and achieved over 65% accuracy for the drowsy class, but attention was raised for the impact that class imbalance and subject-independent strategies can have [10].

There has been work on subject dependent approaches and models agnostic to the sensor acquiring the signal, showing that the same model trained on an ECG sensor can detect drowsiness using HRV derived from other ECG sensors and even Photoplethysmography [11]. However, it was also observed that the provided KSS levels were not reliable, as two individuals may report the same level while only one has his driving impaired. Drivers would tend to wrongly estimate their KSS levels, which resulted in unrealistic progressions for some subjects and numerous misclassifications.

Since HRV monitoring mechanisms are prone to motion and lack of contact artifacts that introduce corrupted data, robustness is a very desirable property. Even though some works [12], [13] study shorter HRV windows and provide insights for drowsiness detection, to the best of our knowledge, the resulting models have not been tested in terms of robustness against faulty input data, either produced by corrupted sensor readings or by malicious attackers. This is the gap in the literature addressed by this work.

## 3    Methods

This section describes the utilized dataset, models, and methodologies. The work relied on the Python 3 programming language and the following libraries: *numpy* and *pandas* for data manipulation, *biosppy* and *pyhrv* for feature computation, *scikit-learn* for the implementation of SVM and KNN, *lightgbm* for LGBM, *xgboost* for XGB, *shap* for feature impact explanations, and *a2pm* for adversarial example generation.

### 3.1    Dataset and Labeling Process

To perform drowsiness detection with ML, it was essential to use a labelled dataset with reliable samples of awake and drowsy signals. The DROZY [14] multimodal dataset, provided by the University of Liege, contains ECG signals, Psychomotor Vigilance Task (PVT) reaction times, and KSS levels of increasingly sleep-deprived individuals. These 14 healthy subjects, 3 males and 11 females, with a mean age of 23 years, performed three sessions: (i) fully awake, (ii) moderately tired, and (iii) sleep deprived.

This dataset was used because the PVT reaction time information it contains enabled the creation of more reliable labels than the self-assigned KSS levels. The PVT values were used in a methodical data labeling process where the first session of each subject was used as a baseline to identify anomalous reaction times in the other two. If most reactions times in a session were significantly higher than the baseline, then the subject was considered drowsy for that entire session. This comparison with the baseline was performed using the Z-Score statistical measurement, with a threshold of 3 standard deviations before a PVT value was considered anomalous.

After the labeling, HRV windows were created with sizes ranging from 60 to 210 seconds. Other sizes were not considered because smaller windows would not provide a long enough signal for the detection and larger windows would be prohibitive in terms of a near real-time analysis. The ECG of each session was split into multiple overlapping windows, and the overlap was established as half the window size. Hence, for a window size of 120 seconds, the last 60 of a sample will be the first 60 of the subsequent one. For a given session, the number of retrieved samples can be defined as:

$$Retrieved\ Samples = \left\lfloor 2\frac{S}{W} - 1 \right\rfloor \qquad (1)$$

where $S$ is the session length in seconds and $W$ is the window size in seconds.

For each resulting sample, a total of 31 HRV features were computed. These included well-established variables like Detrended Fluctuation Analysis (DFA) alpha 1, as well as other possibly relevant variables like the relative powers of High Frequency (HF), Low Frequency (LF), and Very LF (VLF) bands, in percentages of the total power of a sample. The short-term fluctuations for DFA were in the [*4, 16*] interval, and the VLF, LF, and HF bands were in the [*0, 0.04*], [*0.04, 0.15*], and [*0.15, 0.40*] intervals, respectively. Finally, the dataset was randomly split into training and holdout evaluation sets with 70% and 30% of the samples, respectively. To ensure that the sets preserved the same imbalanced class proportions, the split was performed with stratification. Table 1 summarizes the features present in the utilized dataset.

Table 1. Full utilized features and their domains.

| Time | | Frequency | | Nonlinear |
|---|---|---|---|---|
| HR Mean | NN20 | VLF Peak Frequency | HF Peak Frequency | DFA alpha 1 |
| HR Std | pNN20 | VLF Absolute Power | HF Absolute Power | SD1 |
| HR Min | NN50 | VLF Relative Power | HF Relative Power | SD2 |
| HR Max | pNN50 | VLF Logarithmic Power | HF Logarithmic Power | SD2/SD1 Ratio |
| SDSD | SDNN | LF Peak Frequency | Total Power | |
| RMSSD | | LF Absolute Power | LF Normalized Power | |
| | | LF Relative Power | HF Normalized Power | |
| | | LF Logarithmic Power | LF/HF Ratio | |

### 3.2 Models and Feature Selection

Several experiments were performed considering four ML algorithms: Support Vector Machine (SVM) [15], K-Nearest Neighbors (KNN) [16], Extreme Gradient Boosting (XGB) [17], and Light Gradient Boosting Machine (LGBM) [18].

SVM is a binary classifier that attempts to find a hyperplane capable of segregating two classes, whereas KNN determines the class of each sample according to the plurality vote of its nearest neighbors. To ensure that every feature was scaled to the same value range, these two algorithms were combined with Min-Max normalization. XGB performs gradient boosting with an ensemble of decision trees, with a level-wise growth strategy, and it was used with the Histogram method to compute fast approximations to choose the best tree splits. LGBM also performs gradient boosting with tree ensembles, but it relies on a leaf-wise strategy and Gradient-based One-Side Sampling.

These algorithms were used to create distinct models for each window size, after a fine-tuning process with a grid search over well-established hyperparameter combinations. To determine the best combination for each model, a 5-fold cross-validation was performed, and then the models with the best combination were retrained with a complete training set. Shapley Additive Explanations (SHAP) [19] was used to explain the predictions of each model through a game theoretic approach, ranking the features by their impact. The features with highest impact across the best models were selected, and new smaller models were created to be used in a robustness analysis.

### 3.3 Analysis Methodology

To ensure an unbiased analysis of the adversarial robustness of the considered ML algorithms, the methodology introduced in [20] was replicated for drowsiness detection.

Two training approaches were utilized: regular and adversarial training. In the former, the models were directly created with the original training sets, whereas in the latter, the training data was augmented with realistic adversarial examples that considered the correlations between different features. Afterwards, model-specific adversarial sets were generated by evasion attacks, attempting to deceive a model and cause misclassifications from the drowsy class to awake. Since an attacker would not likely have access to a model's training data in a real deployment scenario, the attack was only performed with access to the holdout sets and to the class predictions of each model.

The adversarial examples of drowsiness detection samples were generated using the Adaptative Perturbation Pattern Method (A2PM) [21]. It relies on sequences of adaptative patterns that learn the characteristics of each class and create constrained data perturbations, according to the provided information about the feature set. The patterns record the value intervals of individual features and value combinations of multiple features, which are then used to ensure that the generated examples remain valid within the dataset's structure and coherent with the awake and drowsy classes.

The method was configured to use independent patterns for the awake and drowsy classes, to account for the possible values of each feature and the correlations between multiple features. Regarding the data augmentation for adversarial training, it used a simple function to create a single perturbation in a copy of each drowsy sample of that set. Hence, a model was able to learn not only from a sample, but also from a simple variation of it. This is especially important to ensure that the model behaves in a similar way with both regular samples and adversarial examples [22]. Regarding the adversarial sets, they were the result of full A2PM attacks, which created as many data perturbations as necessary in the holdout sets until every drowsy sample was misclassified as awake or a total of 30 attack iterations were performed.

### 3.4 Evaluation Metrics

To perform a trustworthy evaluation of a model's robustness, its performance on the original holdout set was compared to its performance on the respective adversarial set. For this evaluation, several evaluation metrics were considered [23], [24].

Accuracy is a standard classification metric that measures the proportion of correctly classified samples. However, its bias towards the majority class must not be disregarded when the minority classes are particularly relevant [10], which is the case of the drowsy class in driver drowsiness detection. Therefore, other metrics like precision and recall can be more reliable. Precision measures the proportion of drowsy predictions that are actual drowsy samples, whereas recall measures the proportion of drowsy samples that were correctly identified, reflecting a model's ability to detect drowsy individuals.

These metrics are indirectly consolidated in the F1-Score. A high F1-Score indicates that drowsy samples are being correctly identified and there are low false alarms. Due to its adequacy for imbalanced data, the macro-averaged F1-Score was preferred for both the fine-tuning process and the performance evaluation. It is defined as:

$$F1\text{-}Score = \frac{2 * P * R}{P + R} \qquad (2)$$

where $P$ and $R$ are the precision and recall of a model's predictions.

## 4     Results and Discussion

This section presents the results obtained in the window and algorithm experiments, in the feature impact analysis, and in the adversarial robustness analysis.

### 4.1     Window and Algorithm Experiments

Multiple experiments were performed to assess the viability of the considered window sizes and ML algorithms for the classification of awake and drowsy HRV signals. The highest F1-Score, 81.58%, was reached by the SVM model that processed windows of 120 seconds: SVM 120. Nonetheless, it is pertinent to highlight the reliability of XGB, which exhibited similar results across all window sizes and achieved the second and third best scores, 77.28% by XGB 150 and 77.21% by XGB 90. Despite KNN being more volatile between different sizes, KNN 180 had a very good generalization and was able to reach 77.07%. Regarding LGBM, its scores were generally lower than the remaining models, with a prominent decline in larger sizes (see Fig. 1).

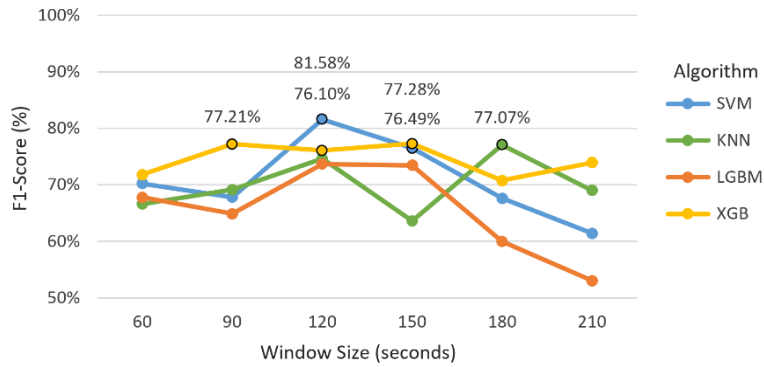

**Fig. 1.** Window and algorithm experimental results.

Overall, the highest results were obtained with windows of 120 and 150 seconds, which suggests that this may be the optimal range for the application of ML to the detection of drowsy individuals. Still, to prevent funneling this work solely into that range and to enable a more thorough experimentation, the six models with F1-Scores above 75% were chosen for the subsequent feature impact analysis: XGB 90, SVM 120, XGB 120, SVM 150, XGB 150, and KNN 180.

### 4.2     Feature Impact Analysis

Feature rankings were obtained from SHAP for the chosen models. The feature that was particularly relevant to the distinction between awake and drowsy individuals across the six best models was the SD2/SD1 ratio, of the nonlinear domain. Nonetheless, most time domain features reached higher Shapley values than the remaining features, so they seem to have a significant impact. This is especially prominent in SVM 120 and XGB 150, the two models with highest F1-Scores (see Fig. 2).

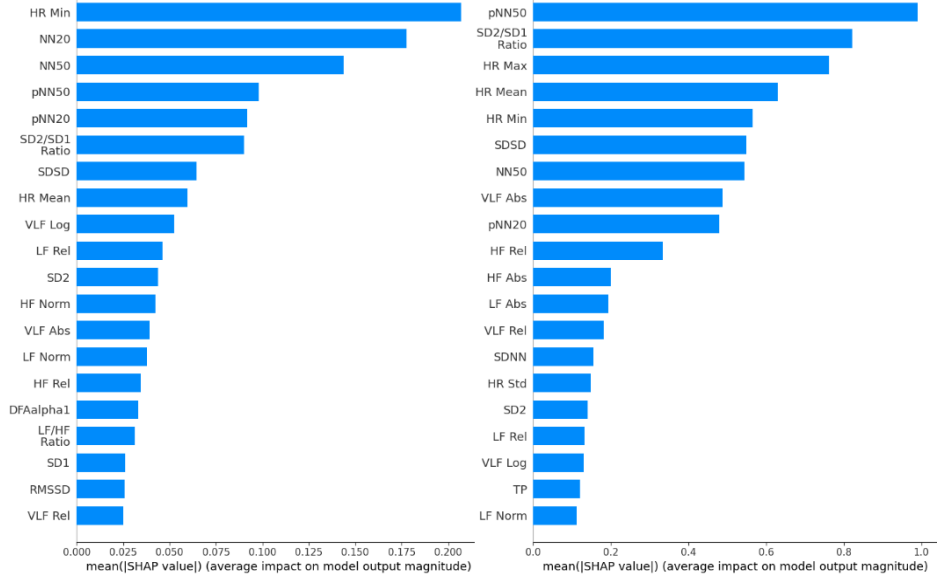

**Fig. 2.** Feature impact for SVM 120 (left) and XGB 150 (right).

Regarding the frequency-domain, it can be observed that the ones with logarithmic and relative powers are generally more impactful than the absolute and even the normalized powers. This suggests that training with relative powers, in percentages of the total power of a sample, can lead to a better generalization and transferability than with the normalized powers. Therefore, the use of relative powers may be beneficial for the application of ML to drowsiness detection with individuals that present significant differences in their HRV signals, such as healthy athletes and smokers.

The most relevant features were selected according to the SHAP rankings for the six models, resulting in 18 numerical features. The new models created with the reduced number of features reached the same F1-Scores on the original holdout set as the initial models with 31 features. Hence, the smaller SVN 120 also reached 81.58%, and the smaller versions of the other five models also reached their respective scores. This indicates that the selected features provide sufficient information for the classification of awake and drowsy samples. Table 2 summarizes the selected features.

**Table 2.** Selected features and their domains.

| Time | | Frequency | | Nonlinear |
|---|---|---|---|---|
| HR Mean | NN20 | VLF Logarithmic Power | HF Relative Power | DFA alpha 1 |
| HR Min | pNN20 | LF Relative Power | LF/HF Ratio | SD2 |
| HR Max | NN50 | | | SD2/SD1 Ratio |
| SDSD | pNN50 | | | |
| | SDNN | | | |

### 4.3   Adversarial Robustness Analysis

An adversarial robustness analysis was performed with A2PM for the smaller models with the selected features, considering both regular and adversarial training approaches. Adversarial examples were generated to attack each model and assess if they could be deceived. Even though all six models created with regular training had high F1-Scores on the original holdout set, numerous misclassifications were caused by the adversarial attacks. Most models exhibited significant performance declines across all windows sizes, with only XGB 90 being able to keep an F1-Score above 60% after the attack. The biggest declines were observed in SVM 150 and SVM 120, which evidences the high susceptibility of SVM to adversarial examples (see Fig. 3).

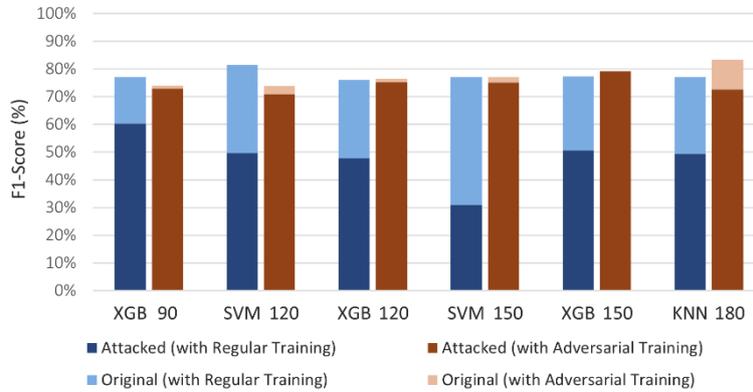

**Fig. 3.** Adversarial attack robustness results.

In contrast, the models created with adversarial training kept significantly higher scores. By training with one realistically generated example per drowsy sample, all models successfully learnt to detect most HRV signal variations. XGB 150 stood out for preserving an F1-Score of 79.19% throughout the entire attack, successfully detecting every generated example. Besides XGB 150, the least affected models were XGB 90 and XGB 120, which highlighted the adversarially robust generalization of XGB.

Regarding the KNN 180 created with adversarial training, even though it was still deceived by a reasonable number of examples and its score dropped to 72.62%, it reached an F1-Score of 83.33% on the original holdout set, a value even higher than the 81.58% previously obtained by SVM 120. Therefore, besides successfully improving the robustness of these models against adversarial examples, the utilized adversarial training approach can also improve their generalization to regular samples.

## 5   Conclusions

This work addressed the use of ML for driver drowsiness detection from feature impact and adversarial robustness perspectives. A total of 31 HRV features of the time, frequency, and nonlinear domains were computed, and PVT reaction times were used in a methodical data labeling process. Then, multiple experiments were performed with different HRV time windows and ML algorithms, the most impactful features were selected, and adversarial evasion attacks were performed with realistic examples against regularly and adversarially trained models to assess their reliability when processing faulty input data and perturbed HRV signals.

The obtained results evidence that the most reliable models in a regular drowsiness detection are SVM and XGB, and the optimal window has between 120 and 150 seconds. Furthermore, the prediction explanations provided by SHAP led to the selection of the 18 most impactful features and to the training of new smaller models. These smaller models achieved the same results as the initial ones with more relative frequency domain features, which indicates that only the selected features are required for the classification of awake and drowsy signals.

Despite their good detection results in regular samples, the adversarial attacks caused significant performance declines, especially in SVM. Nonetheless, XGB stood out for being the least susceptible and preserving high F1-Scores when adversarial training was performed with a function of A2PM, which highlights its robust generalization to faulty input data. Therefore, ML models can significantly benefit from adversarial training with realistic examples to provide a more robust driver drowsiness detection. This is a pertinent research topic to be further explored in the future.

**Data Availability.** A publicly available dataset was utilized in this work. The data can be found at: https://hdl.handle.net/2268/191620.

**Acknowledgments.** This work was done and funded in the scope of the European Union's Horizon 2020 research and innovation program, under project VALU3S (grant agreement no. 876852). This work has also received funding from UIDP/00760/2020.